\pdfoutput=1 
\documentclass[11pt,a4paper]{article}
\usepackage[hyperref]{acl2021}
\usepackage{times}
\usepackage{latexsym}

\usepackage{xurl}
\usepackage{microtype}

\aclfinalcopy 
\def\aclpaperid{2894} 

\title{On the Ethical Limits of Natural Language Processing on Legal Text}

 \author{Dimitrios Tsarapatsanis \\ Law School \\ University of York\\ \texttt{\small dimitrios.tsarapatsanis@york.ac.uk}\\
        \And 
        Nikolaos Aletras \\  Computer Science Department\\ University of Sheffield\\ \texttt{\small n.aletras@sheffield.ac.uk} }

\begin{document}
\maketitle
\begin{abstract}
Natural language processing (NLP) methods for analyzing legal text offer legal scholars and practitioners a range of tools allowing to empirically analyze law on a large scale. However, researchers seem to struggle when it comes to identifying ethical limits to using NLP systems for acquiring genuine insights both about the law and the systems' predictive capacity. In this paper we set out a number of ways in which to think systematically about such issues. We place emphasis on three crucial normative parameters which have, to the best of our knowledge, been underestimated by current debates: (a) the importance of academic freedom, (b) the existence of a wide diversity of legal and ethical norms domestically but even more so internationally  and (c) the threat of moralism in research related to computational law. For each of these three parameters we provide specific recommendations for the legal NLP community. Our discussion is structured around the study of a real-life scenario that has prompted recent debate in the legal NLP research community.
\end{abstract}

\section{Introduction}

Developing computational methods for analyzing legal text is an emerging area in natural language processing (NLP) with various applications such as legal topic classification \cite{Nallapati2008}, court opinion generation \cite{Ye2018} and legal judgment prediction \cite{Aletras2016,Luo2017,Zhong2018,chalkidis-etal-2019-neural}. \emph{Legal NLP} holds the promise of improving access to justice and offers to legal scholars the tools that allow for an empirical analysis of law on a large scale~\citep{Katz2012,zhong-etal-2020-nlp}. 

The development and use of legal text processing technologies also raise a series of ethical questions, on which we focus in this paper. 
For example, following the publication at EMNLP 2019 of a paper on automatic prison term prediction \citep{Chen2019} using a dataset constructed from published and publicly available records of past cases of the Supreme People’s Court of China, a debate ensued about the ethical limits of legal NLP. More specifically, \citet{leins2020} queried in a systematic way whether papers such as that of \citet{Chen2019} should be published. \citet{leins2020} invoked a number of arguments including  considerations to do with the construction of the dataset \citep{Bender2018}, and so-called `dual use' arguments~\citep{Radford2019}, i.e. the possibility of using a system developed for some purpose for another, potentially harmful, purpose. Following a rich discussion, \citet{leins2020} asked whether it is ethically permissible that legal NLP should be used at all to predict items such as prison terms. 

We believe that the kind of ethical query put forth by \citet{leins2020} is vital for the future of legal NLP and computational law in general. However, we also contend that it is essential that a more general discussion should be conducted about the pertinent normative principles and concepts at play. References to more general principles can often curb the temptation to make decisions on the basis of ad hoc moral intuitions which might not be and probably are not, as we explain later on, universally shared. The contributions of this paper are as follows: 

\begin{itemize}
\item In this paper we make no claim to cover all the ethical ground in a comprehensive way. Instead, we focus on three crucial ethical parameters that, to the best of our knowledge, have not been extensively debated so far: (a) the role of \emph{academic freedom} (\S \ref{sec:academic_freedom}); (b) the existence of a wide \emph{diversity of legal and ethical norms} applicable to or endorsed by the global NLP scholarly community (\S \ref{sec:diversity}); and (c) the \emph{threat of moralism} in legal NLP research (\S \ref{sec:moralism}); 
\item We illustrate the practical difference that the first two factors would make, when taken into account, on the basis of the study of real-case scenarios of developing or using legal NLP technology;
\item Moreover, for each of the three normative parameters (i.e. academic freedom, norms diversity and moralism), we provide specific recommendations for the legal NLP community.
\end{itemize}

\section{Academic Freedom}
\label{sec:academic_freedom}

\subsection{What is academic freedom?} 
The idea of academic freedom has a long and important pedigree in the history of the Western university \citep{Newman1976,Searle1972}. Still, despite the importance of the idea, there is currently no commonly accepted definition accepted across domestic jurisdictions and scholars \citep{barendt2010academic}. Be that as it may, for the purposes of this paper we provide the following working definition: \emph{academic freedom is the freedom of scholars, whether employed by universities or not, to decide without undue external pressure or coercion the topics of their research, the standards of such research and the application of such standards to the scholars’ peers}. 

Academic freedom is both similar to and different from the more general right to freedom of speech \citep{barendt2010academic}. It is similar in that it involves communicating freely chosen and conducted research to pertinent audiences. It is different in that: (a) it also applies to non-expressive research activities; (b) it is more circumscribed because scholars are bound by certain professional standards; and (c) it has both an individual-rights dimension and an institutional dimension, both being bound to the idea of the university as an intellectual space of free and uncoerced pursuit of truth. Academic freedom is recognized as a distinctive right by a number of national constitutions, such as Article 5(3) of the German Basic Law, as well as by some transnational legal instruments, such as Article 13 of the EU Charter of Fundamental Rights.

\subsection{What is the value of academic freedom?} 
Two different tacks could be taken. First, one could provide a consequentialist argument. Consequentialist arguments identify right and wrong actions solely on the basis of their good or bad consequences \citep{Driver2012}. Accordingly, academic freedom could be justified by invoking the beneficial results of research undertaken, conducted and debated freely by researchers. 

Second, one could opt for a deontological argument. Deontological ethical arguments identify permissible actions on the basis of (absolute) rules and irrespective of the consequences that actions may bring about \citep{Alexander2007}. Under a deontological construal, the guarantees of academic freedom would not be justified solely by the (expected) outcomes of the research actions they would allow but also: (a) because the pursuit of truth and knowledge for their own sake is a good in itself and irrespective of whether its (expected) consequences are good or bad (under some criterion of `goodness' and `badness'); and (b) because preserving academic freedom also preserves the integrity and inviolability of the very person of the researcher \citep{Nagel1995}.

We hasten to add for the sake of completeness that one could perhaps also opt for a virtue ethical approach to academic freedom~\citep{Hursthouse2018}. We refrain from doing so in this paper because, to the best of our knowledge, no comprehensive literature on a virtue ethics approach to academic freedom exists. Therefore, whatever the merits of such an approach, it falls outside the scope of this paper, which focuses on more entrenched within the pertinent research community ethical parameters.

\subsection{Academic freedom and  legal NLP} 

One first thing to note is that, despite its importance, academic freedom does not seem to be recognized as a distinctive ethical value in many codes related to the community of computer science researchers. For example, the ACM Code of Ethics refrains from specifically referring to that freedom in its first section, which is devoted to `General Ethical Principles'.\footnote{ACM Code of Ethics -- \url{https://www.acm.org/code-of-ethics}} While explaining this gap is beyond the scope of the present paper, we contend that academic freedom should be taken as seriously as the other ethical principles and values invoked therein, especially if one is to accept, as we think one should, that academic freedom's scope of application does not just include the `academia' in the narrow institutional sense (i.e. universities or other kinds of tertiary education institutions) but virtually anyone who engages in research and scholarship with a view to descovering the truth about a subject matter.  

If we begin with the widespread assumption that academic freedom is indeed a fundamental right, certain things seem to follow. First, in the assessment of the ethics of legal NLP, academic freedom will have to be taken into due account and then balanced against other values, such as the value of privacy of data subjects or the value or disvalue of certain further potential applications of systems (e.g. for legal judgment prediction) than those for which they were initially developed (`dual-use'). 
When values are balanced against each other, the all-things-considered permissibility of taking an action (in our case proceeding with a research project or publishing a paper in legal NLP) ultimately depends on taking all relevant contextual factors into account. We thus contend that there is no automatic general rule to apply to all cases. Still, we believe that, for example, in most imaginable scenarios where there is minimal interference with the rights of data subjects insofar as data such as court judgments are harvested from the public domain, pursuing an otherwise cognitively valuable research project should be permitted. More generally, taking academic freedom seriously renders the ethical permissibility of legal NLP projects more complex than sometimes acknowledged, since academic freedom will have to be taken seriously into account as an independent ethical factor, something which rarely (if ever) seems to happen as things currently stand, and despite the fact that almost all researchers pledge some kind of commitment to some version of academic freedom. 

Second, and more importantly, ethical assessments of legal NLP research will also depend on the particular interpretation of the value of academic freedom \citep{barendt2010academic}. For example, deontological interpretations of that value radically circumscribe the permissibility of invoking certain kinds of consequentialist reasons to block particular types of legal NLP research \citep{Waldron2000}. Invoking such reasons, we contend, is a direct violation of the integrity and equal freedom of researchers in the same way that, say, invoking people’s tendency to be offended by certain kinds of artistic creations (for example, `blasphemous' art) to block dissemination of art is a violation of the dignity of the artist. At the very least, under a robust understanding of the right to academic freedom \citep{Dworkin1977}, arguments to the effect that a certain piece of legal NLP research will have `bad' consequences (whatever these are) will not be enough unless the risk of the consequences actually (as opposed to merely speculatively) obtaining is clear, present and significant. Again, we stress that there is no easily applicable general ethical rule here. Researchers should apply their ethical judgment in a contextual way by trying to account for all the relevant factors. We provide in the discussion that follows a concrete illustration of what a deontological understanding of academic freedom might entail in a particular scenario.

We also argue that even consequentialist interpretations of the value of academic freedom will place emphasis on the benefits of allowing researchers, at least \emph{prima facie}, to freely conduct research, even when such research is ethically controversial. These benefits consist, among other things, in the possibility of maximizing the chances of getting to the truth about a certain subject-matter (in our case, law) through the proliferation of different research perspectives, thus promoting the public good.\footnote{ACM Code of Ethics -- \url{https://www.acm.org/code-of-ethics\#h-1.2-avoid-harm}} Moreover, in this case again, the merely speculative possibility of certain adverse effects obtaining (e.g. systems designed by the researchers being used by others to pursue unethical goals) is not enough to render such research unethical in itself. There must be a real probability, not just a theoretical possibility, that these effects might in fact materialize.

\subsection{Specific scenarios} 
These points can be readily illustrated by reference to the article by \citet{leins2020} with which we began our discussion. We have said that, at the very least, academic freedom should be taken into account as an independent ethical value. We contend that this would modify significantly at least the structure of the argument provided by \citet{leins2020}. We stress, of course, that this in no way implies that academic freedom `wins by default'. In fact, we have already highlighted that academic freedom is not the only pertinent ethical value and we have stressed that we do not believe that a generally applicable rule exists. All we intend to do is to argue that academic freedom as such should be taken seriously into account and balanced against other pertinent ethical factors. However, we think that in the context of the particular case under consideration, academic freedom tips the balance in favor of the ethical permissibility of the research by \citet{Chen2019}. Thus, the `dual-use' and `dataset construction' concerns voiced by \citet{leins2020} will not be sufficient on their own to call the shots in ethics assessment in favor of rejecting either conducting or publishing the scrutinized research for the following reasons. 

\paragraph{Dual-use}
Regarding the `dual-use' concerns, we think that \citet{leins2020} overestimate the dangers of an algorithm designed by academics being used to decide real cases with adverse consequences for real people. In particular, \citet{leins2020} provide no reason to worry that any such use might happen anytime soon, nor evidence that there is, for example, a serious standing intention on the part of Chinese authorities to implement what would amount to a radical reform of the judicial system. Accordingly, the chance of an ethically unacceptable use of the designed algorithm seems, on the face of it, rather small (not to say almost completely theoretical). If this should be accepted, then both deontological and consequentialist interpretations of academic freedom tip the balance in favor of conducting and publishing the scrutinized research. Deontological interpretations place a very high premium on curtailments of the freedom and dignity of researchers. Arguably, this premium is not met by just the theoretical possibility of some bad consequence eventually resulting somewhere downstream from the research. But even consequentialist interpretations of academic freedom would concur in the case at point. This is because the expected probability of the occurrence of the bad consequences highlighted in the paper by \citet{leins2020} (use of a potentially biased algorithm by the judiciary to help make inequitable decisions) seems in this case particularly low (or perhaps even nonexistent). 

Of course, we fully acknowledge that we might be wrong in our assessment about the probability of a potentially evil `dual-use'; we are in no way experts about Chinese legal and political affairs or of impending developments in the Chinese legal system. Still, the reasoning we have outlined also has a dimension to do with accruing and assessing evidence for ethical assessment. Thus, we hold that, in cases of doubt or where adequate evidence is neither available nor forthcoming, the value of academic freedom grounds at least a presumption in favor of proceeding with the research, the burden of proof in `dual-use' scenarios been shifted to those who believe that there is a serious reason not to undertake a given piece of research. It is crucial for academic freedom that this burden should not be met by the researchers themselves. Arguably, in the case at hand the evidential burden has not been met by \citet{leins2020}.        

\paragraph{Dataset Construction}
Similar considerations apply when it comes to assessing concerns to do with the construction of any particular dataset for legal NLP. The argument put forth by \citet{leins2020} stresses that the dataset exposes people to harm because the defendants of past cases are identifiable. But arguably the probability of exposing people to harm through the construction of a dataset consisting of already decided cases by some domestic Supreme Court seems so low as to be practically non-existent. On the one hand, \citet{leins2020} do not provide any evidence in favor of their argument. On the other hand, and to the best of our knowledge, the probability of these cases reopening together with that of some judge being specifically influenced by the dataset if such reopening occurs is practically non-existent. In all jurisdictions that we know about, cases by courts at the top of the judicial hierarchy are mostly, even if not uniquely, to do with the past, since by definition Supreme Courts issue the final judgment on some case. Thus, in order for considerations of potential harm to defendants to trump academic freedom, the risk of harm must be real, even under a generous understanding of the latter term, not merely imaginary or speculative. We stress, moreover, that this kind of understanding of harm coheres well with the ACM Code of Ethics that we have already cited. Thus, here again academic freedom, under any reasonable interpretation, should normally tip the balance in favor of proceeding with the research. 

In a similar way, contending that some dataset `unfairly advantages or disadvantages’, as \citet{leins2020} claim, appears to hugely overstate the real degree to which academic research might impact on the actual workings of judicial and political institutions. This is even more the case with respect to the specific dataset in question, since the latter was created by using documents already available in the public domain. In fact, it is the availability of the documents in the public domain itself, for which researchers bear no ethical responsibility, that creates certain risks to data subjects, if at all. Furthermore, such a contention bypasses the obvious fact that the dataset merely indicates the existence of unfair treatment that happened, if it did, at the level of what some Supreme Court did and not at the level of converting the activity of this Court into a dataset. 

Last, but not least, here again the cases refer to facts that have already happened and no details are provided in \citet{leins2020} as to whether there is, within the Chinese legal and social order, a concrete and standing threat for individuals to be unfairly treated on the basis of past cases (whether these led to convictions or not). Thus, we repeat that a merely theoretical or speculative possibility should not be considered sufficient, absent other factors, and especially concrete and real, not just imagined, reasons to trump academic freedom.

\subsection{Recommendations}
\emph{In decisions about the ethics of legal NLP research, deliberation should begin by commencing with a (rebuttable) presumption to the effect that academic freedom should not be curtailed lest there be compelling, to wit, clear and present or at least significantly probable (not purely theoretical or speculative) reasons to decide otherwise}.

\section{Diversity of Ethical and Legal Norms}
\label{sec:diversity}

\subsection{Norms Diversity: The general problem}

The community of legal NLP researchers is now global, ranging from the Global North to the Global South. However, a number of ethical standards on how to conduct legal NLP research in many cases seem to be either local or of local origin. Moreover, different ethical and legal standards may be found across different jurisdictions or cultures. Moral diversity is thus an issue that crops up in a variety of contexts to do with legal NLP. Below, we shall concentrate on the specific issue of privacy and data protection in the construction of datasets. Still, similar issues could also arise in other areas, such as the definition of what counts as `unfairness' towards a specific group. 

Privileging a particular ethical or legal standard over another (or a particular interpretation of a common standard) could fuel the suspicion that, instead of reflecting a perspective of detached and `pure rationality', such standards and interpretations are in fact just entrenching local (and, for that matter, Western European or, more generally, Global Northern) prejudices and preconceptions, imposing them as mandatory norms on researchers who might not reflectively endorse them. There are two distinct but interrelated issues here. First, as a matter of fact, there appears to be a wide diversity of ethical standards that are accepted by different communities, with some of them being endorsed by some and rejected by others and vice versa \citep{Prinz2007}. Second, even when ethical or legal standards are shared across communities, there is often more or less widespread and reasonable disagreement about their `best' or `proper' interpretation \citep{Gowans2015}. 

Now, the above facts do not immediately lead to what might seem like an unacceptable ethical relativism, i.e. the idea that no universal valid ethical standards could ever exist \citep{Gowans2015}. A discussion about the merits of ethical relativism is outside the scope of this paper, and we wish neither to endorse nor to criticize it. Our only point here is that, irrespective of the stand one takes on the issue of ethical relativism, the reality of diversity of ethical and legal opinion gives rise to an issue that cannot be avoided, to wit, the kind of position that researchers may take when confronted with research based on ethical rules and norms that significantly diverge from their own. Thus, ethical diversity points to the need to submit the content of the researcher's own standards, especially when they are used to evaluate the research of scholars who do not necessarily endorse them or reside in parts of the globe where they are considered neither legally nor perhaps even morally binding, to a much more searching examination. Accordingly, it should not be taken for granted that, say, historically and geographically contingent conceptions of a particular concept (e.g. data privacy in the EU, or the ban of judge analytics in France\footnote{\url{https://www.artificiallawyer.com/2019/06/04/france-bans-judge-analytics-5-years-in-prison-for-rule-breakers/}}) should always and without further argument take normative priority in ethical assessments.

Like in the previous section on academic freedom, we hasten to add that this does not amount to an easily applicable rule or `formula' (which in our case could be either `pay no heed to alien rules and norms and stick to what you think is right' or its exact opposite such as `accept diversity no matter how unethical the alien rules or norms appear to be'). In fact, we do not think that there exists any easy, uncontroversial and unequivocal way by which to resolve such ethically complicated issues. As everywhere in life, a lot will depend on the particulars of each case and on the researcher's capacity to manifest a sensitivity to these particulars and of the specific ways in which they mesh in each scenario. All that we are suggesting is that issues to do with diversity of ethical and legal outlooks should be brought to the attention of researchers and, in the end, may only be resolved by the exercise of contextual and inherently controversial ethical judgment. At the very least, the actual presence of a diversity of different outlooks requires a willingness to engage seriously with 'alien' rules and norms, trying to understand their point and justification from the other's point of view, instead of easily and sometimes even complacently dismissing them out of hand as straightforwardly unethical. 

\subsection{Specific Scenario}

To better apprehend how such a relatively abstract discussion may play out in a particular context and scenario, let us once again revisit the initial example with which we began our discussion, i.e. the paper by \citet{leins2020}. 

\paragraph{Data Privacy: The ethical factors at play}
A large part of the argument put forth by \citet{leins2020} consists in querying the process of construction of the dataset used by \citet{Chen2019} by invoking the rights of data subjects to privacy (similarly, \citet{Aletras2016}, \citet{chalkidis-etal-2019-neural}, \citet{chalkidis-etal-2020-legal} and \citet{chalkidis-etal-2021-paragraph} use publicly available data that contain private information in the context of the European Court of Human Rights). Thus, \citet{leins2020} raise issues to do with the use of `sensitive and confidential data', such as data to do with prison time served. They also refer to the fact that the people represented in the dataset do not seem to have been informed about data collection. The question then is whether these research practices are ethically problematic. While it seems self-evident, at least for a researcher accustomed to the practices of late twentieth century European states, to answer in the affirmative, an issue that \citet{leins2020} do not address is that any answer to this question will presuppose a particular prior understanding of the content of the right of data subjects to privacy, an understanding, moreover, that may only be available to certain kinds of legal and ethical cultures and not to others.  

The discussion may progress by reference to the simple observation that, when it comes to the legal and ethical status of datasets composed of data already available in the public domain (such as those in the papers by \citet{Chen2019} and \citet{chalkidis-etal-2019-neural}), there appears to be no global universal consensus as to the scope of the privacy rights of the data subjects involved.\footnote{\url{ https://www.cnil.fr/en/data-protection-around-the-world}} A few jurisdictions deploy very demanding definitions of data privacy, coupled with special protection provided with respect to 'sensitive data': the EU and Australia are prime examples. Still, there also exist major divergences between state practices, some states being much more permissive in those respects than others. 

The invocation of privacy as an ethical consideration thus raises a distinctive problem: how should privacy be understood in view of the fact that there is no universal consensus on its nature and its protection across states? Should particular definitions of data privacy constitute the appropriate normative benchmark? And why? As we have already said, some of these definitions might be significantly more demanding than others, reflecting different local conventions about the value of rights that data subjects have under the law. 

We contend that the documented divergences in the attitudes towards data protection and the right to privacy, differences reflected in the law of different jurisdictions, in fact stem from a more fundamental phenomenon, i.e. the diversity of ethical, moral and legal outlooks both across time and across states, cultures and peoples. \citep{Prinz2007}. In fact, as we have already shown, it is arguable that the high standard of data protection that \citet{leins2020} use, which also appears to be the standard used by the ACM Code of Ethics\footnote{\url{https://www.acm.org/code-of-ethics}} is in the minority from a global point of view \citep{Renteln1988}. Moreover, the construction of datasets from databases containing publicly available court judgments appears to at least count in favor of the prima facie permissibility of that research practice. 

The availability of court judgments in the public domain even in core Western countries that otherwise protect private data in other domains and with respect to international courts such as the European Court of Human Rights still belies an older conception of privacy: one in which the publicity of court trials as a rule of law guarantee trumps the individual’s wish to hide oneself from others in public space \citep{Langford2009}. Under this older conception of privacy, not only is it possible to retrieve the names of litigants in past cases, but also that possibility is protective of the rule of law in the same way that, say, secret trials of terrorist or other suspects are still widely thought to undermine it \citep{Resnik2011}. The above comments thus suffice to show that the high standard of privacy protection that is sometimes used to scrutinize research practices in legal NLP, exemplified by \citet{leins2020}, is not only geographically but also historically contingent.

Under these conditions, the following question becomes ethically pressing: Why should any local and geographically restricted conception of stricter data privacy and protection be preferred to, say, a more relaxed and minimal one? Granted, researchers conducting legal NLP research in locations where that particular conception is prevalent and legally obligatory, such as the European Union or Australia, should of course abide by it; but why should other researchers, such as those residing, say, in Singapore, the United States or Africa, be held to that same particularly demanding standard? Moreover, is it even fair and ethical, given what was already stated above with respect to academic freedom, to hold researchers (e.g. non-European and non-Australian) to a peculiarly local, and by no means universally acknowledged, standard?

\paragraph{Data Privacy: Application to the case at hand} 
On the basis of the above considerations, we believe that this kind of contextually informed thinking could argue in favor of the ethical acceptability of the research practices used by \citet{Chen2019} in the construction of their dataset. There are two important variables at play here. On the one hand, the data is already in the public domain. Under these circumstances, the interference of the researchers with the data subjects' rights appears  minimal at worst: in fact, even under a demanding definition of privacy, the main interference with privacy rights is due to the availability of the publicly available database itself and not to the researcher's activity of constructing a dataset for research purposes out of it. But, second, we could even go further and ask why what appears like a specifically European definition of privacy and data protection, and moreover one which, at least as things stand nowadays  seems to be more of an outlier on a global level, should be adopted as a normative yardstick by researchers working in a completely different ethical and regulatory environment. 

As we have already said, there is a long and venerable Western and European tradition which lays stress on the importance of the publicity of court judgments. For a long time, it was considered not unethical but utterly normal to make court trials and their outcomes public, so as to control through public scrutiny the exercise of state coercion on individuals. While this consideration might now seem dated in some parts of the world, especially those that have entrenched traditions of judicial independence and a high degree of trust to the judicial system, we should be sensitive to its importance in other parts of the world. It thus becomes difficult to resist the conclusion that the absence of respect of (putatively European) data protection norms by  \citet{Chen2019}) is not a fatal ethical objection to either conducting or publishing their research.

\subsection{Recommendations}
\emph{Forging genuine universal ethical standards requires a global conversation between researchers engaged from a plurality of standpoints and traditions. When such standards do not exist or exist only to a minimal degree, ethical assessment for global conferences, journals and reviews should be appropriately flexible and respectful of differences and reasonable disagreements. No automatic assumptions should be made that a `one-size-fits-all' model is sufficient to make informed decisions about the ethical status of research practices}.

\section{The Threat of Moralism in Legal NLP}
\label{sec:moralism}
We believe that the previous discussion points to a more general issue, that we treat under the heading of the `threat of moralism’ in legal NLP. While this issue is more theoretical than the previous ones (which had practical ramifications), we think it should be treated on its own, insofar as it allows us to propose a more general and comprehensive conceptual framework that allows us to make wider sense of the ethical issues at play. This section of our paper thus has the function of both subsuming the previous discussion and further advancing it, albeit without this time focusing on the discussion of a specific scenario.

\subsection{How to characterize moralism?}
Moralism can be intuitively understood as `the vice of overdoing morality' \cite{Coady2005}. In the context of qualitative research, \citet{Hammersley2011} has contended that moralism can take two different forms. First, it might involve the belief that substantive ethical values, other than the disinterested pursuit of knowledge for its own sake, should be integral goals of research. Second, it might involve the requirement that researchers adhere to `high' or even the `highest possible' ethical standards \citep{Hammersley2011}. 

Here, and with no ambition of a comprehensive discussion, we roughly define moralism as the idea that substantive moral values and constraints more demanding than mere adherence to valid legal norms or to relatively uncontroversial and in any event minimal ethical norms, such as the requirement not to harm others, must be taken into account when assessing research outputs, including legal NLP research. Our argument is that moralism of this sort threatens academic freedom and the equal dignity of researchers as bearers of such freedom.

\subsection{Why moralism might be a problem in legal NLP?}
To think clearly about why moralism in the sense of the pursuit of substantive moral values might be a problem in legal NLP, we might make an analogy with moralism in the setting of ordinary social and political life. At least liberal democracies take seriously John Stuart Mill's idea \citep{Mill2015} that people should be free to pursue various ends that they themselves set, so long as they keep within certain reasonable limits circumscribed by the so-called `harm' principle, i.e. the requirement not to harm others equally engaged in the pursuit of their proper ends. 

A political community that attempts to impose substantive ends on individuals, say on the assumption that certain forms of life are `higher' or `more important' than others is guilty of disrespecting the autonomy of individuals. We contend that, in an analogous way, a research community that attempts to instill in its members specific ethical ends other than the disinterested pursuit of truth and knowledge, especially by ethically assessing the very content of research endeavors, is risking falling into a kind of moralism which fails to take the freedom of researchers seriously. In particular, the mere fact that certain members of the research community subjectively find a piece of research to be unacceptable on other than professionally defined and accepted methodological grounds or basic and flagrant disrespect of moral norms such as the no-harm principle, should not be deemed automatically sufficient for ethical condemnation of the research. 

\subsection{Recommendations}
\emph{The primary moral duty of legal NLP researchers, like all researchers, is to the disinterested pursuit of truth as they understand it, and not to substantive ends which are extrinsic to that pursuit}.

\section{Conclusions}

In this paper we have contributed to the ongoing discussion on the ethics of legal NLP~\citep{leins2020}. We laid emphasis on three normative factors whose importance, to the best of our knowledge, has not been sufficiently acknowledged, i.e. academic freedom, diversity of ethical and legal norms of the global NLP community and the wider and more abstract issue of moralism in research ethics. Moreover, we illustrated how the first two factors might make a practical difference to ethical decision-making by a detailed discussion of a specific scenario. We also stressed that these factors do not amount to any automatically applicable general rule but require the exercise of contextual ethical thinking on the part of researchers. Final decisions should be made by taking all ethically relevant factors into account. 

We believe that the ethical factors we identified can help the legal NLP community become more reflective and tolerant of a wider variety of approaches whilst at the same time remaining fully committed to the academic ideal of disinterested pursuit of truth for its own sake.

\bibliography{anthology,custom}
\bibliographystyle{acl_natbib}

\end{document}